\documentclass[]{IEEEtran}
\ifCLASSINFOpdf
\else
\fi
\usepackage{amsmath,amssymb,amsfonts}
\usepackage{algorithmic}
\usepackage[ruled,vlined,linesnumbered]{algorithm2e}
\usepackage{algorithmic,algorithm2e,float}
\SetKwInput{KwInput}{Input}                
\SetKwInput{KwOutput}{Output}              
\SetKwInput{KwRepeat}{Repeat}
\SetKwInput{KwUntil}{Until}
\SetKwInput{KwReturn}{Return}
\usepackage{amsfonts}
\usepackage{graphicx}
\usepackage{textcomp}
\usepackage{xcolor}
\usepackage{tabularx}
\usepackage{array}
\usepackage{booktabs}
\usepackage{threeparttable}  
\usepackage{multirow}
\usepackage{bm}
\usepackage{amsmath}
\usepackage{subfigure}
\usepackage{amssymb}
\usepackage{pifont}
\newcommand\drop[1]{{\color{black}#1}}

\usepackage{adjustbox} 
\begin{document}
%
\title{EmT: A Novel Transformer for Generalized Cross-subject EEG Emotion Recognition}
%
%
%

\author{Yi~Ding,~\IEEEmembership{Member,~IEEE,}
Chengxuan~Tong,~\IEEEmembership{Graduate Student Member,~IEEE,}
Shuailei~Zhang,
Muyun~Jiang,
Yong~Li,
Kevin~Lim~Jun~Liang,
and~Cuntai~Guan,~\IEEEmembership{Fellow,~IEEE}
\thanks{Yi Ding and Chengxuan Tong contribute equally to this work.}
\thanks{Yi Ding, Chengxuan Tong, Shuailei Zhang, Muyun Jiang, and Cuntai Guan are with the College of Computing and Data Science, Nanyang Technological University, 50 Nanyang Avenue, Singapore, 639798. E-mail: (ding.yi, tong0110, shuailei.zhang, james.jiang, ctguan)@ntu.edu.sg.}
\thanks{Chengxuan Tong and Jun Liang Kevin Lim are with Wilmar International, Singapore.
E-mail: (chengxuan.tong, kevin.limjunliang)@sg.wilmar-intl.com.}

\thanks{Yong Li is with the School of Computer Science and Engineering and the Key Laboratory of New Generation Artificial Intelligence Technology and Its Interdisciplinary Applications, Southeast University, Nanjing 210096, China. Email: mysee1989@gmail.com

}
\thanks{Cuntai Guan is the Corresponding Author.}
}

%
%

\markboth{IEEE Transactions on Neural Networks and Learning Systems}%
{Shell \MakeLowercase{\textit{et al.}}: Bare Demo of IEEEtran.cls for IEEE Journals}
%



\maketitle

\begin{abstract}
Integrating prior knowledge of neurophysiology into neural network architecture enhances the performance of emotion decoding. While numerous techniques emphasize learning spatial and short-term temporal patterns, there has been limited emphasis on capturing the vital long-term contextual information associated with emotional cognitive processes. In order to address this discrepancy, we introduce a novel transformer model called emotion transformer (EmT). EmT is designed to excel in both generalized cross-subject EEG emotion classification and regression tasks. In EmT, EEG signals are transformed into a temporal graph format, creating a sequence of EEG feature graphs using a temporal graph construction module (TGC). A novel residual multi-view pyramid GCN module (RMPG) is then proposed to learn dynamic graph representations for each EEG feature graph within the series, and the learned representations of each graph are fused into one token. Furthermore, we design a temporal contextual transformer module (TCT) with two types of token mixers to learn the temporal contextual information. Finally, the task-specific output module (TSO) generates the desired outputs. Experiments on four publicly available datasets show that EmT achieves higher results than the baseline methods for both EEG emotion classification and regression tasks. The code is available at \textit{https://github.com/yi-ding-cs/EmT}.
\end{abstract}


%
\IEEEpeerreviewmaketitle
\section{Introduction}
\IEEEPARstart{E}{motion} recognition using electroencephalography (EEG) plays an important role in brain-computer interface (BCI) assisted mental disorder regulation. It requires machine to perceive human emotional states from brain activities using artificial intelligence techniques \cite{7946165,8634938}. The predictions of emotions can be used for neurofeedback in the regulation process \cite{ehrlich2019closed}. Accurate predictions and fair generalization abilities to unseen subjects are crucial in building robust real-world BCI systems. Deep learning methods have shown promising results for accurate detection of brain activities \cite{doi:10.1080/27706710.2023.2181102,9508768}. The design of the neural network architecture becomes crucial. 

Incorporating neurophysiological prior knowledge into neural network architectures improves emotion decoding performance \cite{ijcai2018p216,9091308,9762054}. Common considerations include left-right hemisphere asymmetries, neurophysiologically meaningful graph connections, and the temporal dynamics of EEG signals. According to neuropsychological studies, the left and right hemispheres react differently to emotions, particularly in the frontal areas \cite{coan2004frontal}. Several deep learning methods \cite{9762054,li2018novel,huang2021differences} draw inspiration from this prior knowledge, achieving improved emotion decoding performance. EEG electrodes are placed on the scalp, which naturally forms a non-Euclidean structure. Therefore, many studies treat EEG signals as graphs, using either a neurophysiologically designed adjacency matrix \cite{9091308} or a learnable one \cite{8320798,8815811}. However,  there is still more prior knowledge that should be investigated. As one of the high-order cognitive processes in the brain, emotion consists of more basic processes such as attentional, perceptual, and mnemonic system processes \cite{KOBER2008998}. Different brain regions are cooperatively activated under different cognitive processes, e.g., frontal and parietal networks in attention \cite{doi:10.1152/jn.01052.2004} and medial temporal lobes, prefrontal cortex, and parietal cortex interactions for episodic memory \cite{watrous2013frequency}. Using a pre-defined or a single learnable adjacency matrix cannot capture complex brain region connectives underlying emotional processes. Another prior knowledge about emotions is that emotional states are continuous in short periods while not consistent along the long stimuli \cite{9698041}. Less attention is paid to this temporal contextual information underlying emotions. 

\begin{figure}[t]
    \centering    \includegraphics[width=1\linewidth]{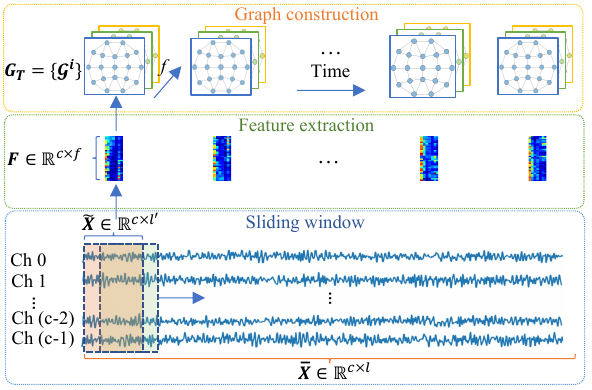}
    \caption{Illustration of TGC. Each segment, $\bar{X}$, is split into several sub-segment, $\tilde{X}$. Features in different frequency bands are extracted for each $\tilde{X}$ channel by channel to form $\boldsymbol{F}$. Then each EEG channel is regarded as a node, and the extracted features are treated as node attributes. Combing all the graphs which are in time order, we get the temporal graphs, $\boldsymbol{G}_{T}$.}
    \label{fig:temporal_graph}
\end{figure}

To address the above-mentioned problems, we propose a novel transformer-based structure, named emotion transformer (EmT), for both generalized cross-subject emotion classification and regression tasks. To the best of our knowledge, this is the first work to explore transformer-based structures on cross-subject EEG emotion classification and regression tasks together. To learn temporal contextual information, we represent EEG segments as temporal graphs, as shown in Figure~\ref{fig:temporal_graph}. We learn the spatial information of each EEG graph to form a token. Then the long-short time contextual information is learned upon the token sequence. 

We propose a residual multi-view pyramid graph convolutional neural networks (GCN) module, named RMPG, to capture multiple EEG channel connections for different basic cognitive processes in emotions. Multiple GCNs with independent learnable adjacency matrices are utilized parallelly in RMPG. Different from \cite{Song_Liu_Zheng_Zong_Cui_2020} that uses multiple adjacency matrices among manually defined local groups, we learn the connections among all the EEG channels. The reasons are two-fold: 1) locally defined groups are special cases of a global connection in which the EEG channels are fully connected within each local group, and 2) more specific channel-wise connections can be learned. Those parallel GCNs have different numbers of layers that can learn multi-view and multi-level graph embeddings with the help of their learnable adjacency matrices. Together with the output of a residual linear projection branch, a feature pyramid is formed. A mean fusion is utilized to combine the information in the feature pyramid as one token. Hence, a temporal sequence of tokens is formed.

\begin{figure*}[t]
    \centering
    \includegraphics[width= 0.98\linewidth]{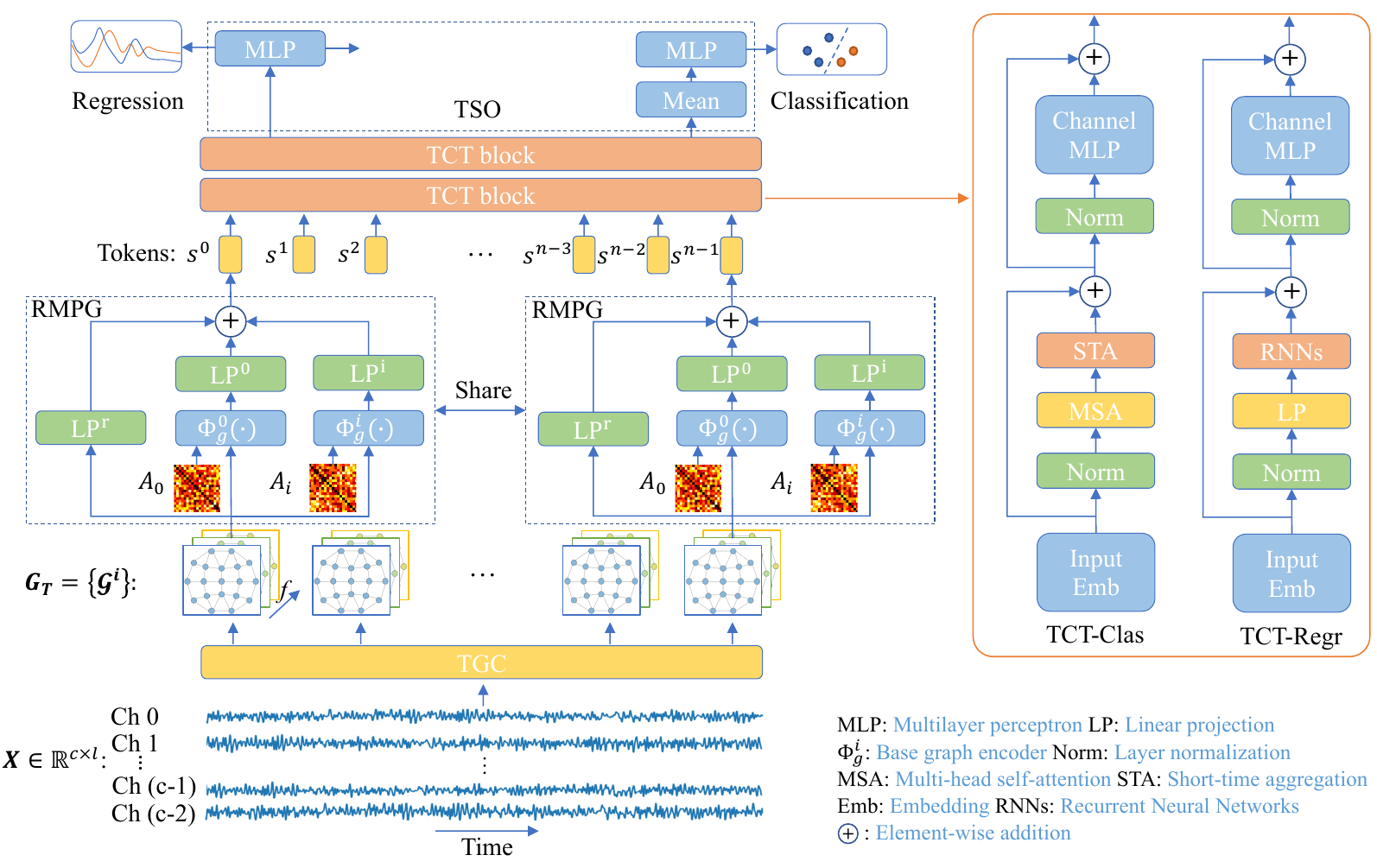}
    \caption{The network structure of EmT. The temporal graphs from TGC are used as the input to RMPG that will transfer each graph into one token embedding. Then TCT extract the temporal contextual information via specially designed token mixers. We propose two types of TCT structures, named TCT-Clas and TCT-Regr, for classification and regression tasks separately. A mean fusion is applied before feeding the learned embeddings into MLP head for the classification output. For regression tasks, a MLP head projects each embedding in the sequence into a scalar to generate a sequence that can be used to regress the temporally continuous labels.}
    \label{fig:EmT}
\end{figure*}

We further propose temporal contextual transformers (TCT) with different token mixers \cite{Yu_2022_CVPR} to learn contextual information from the token sequence for EEG emotion classification and regression tasks. An EEG trial refers to the period when one type of stimuli, e.g., a happy movie clip, is presented to the subject while the EEG is recorded. The label of the entire trial is assigned as happy according to either the self-assessments or the stimulus contents in classification tasks. However, the emotional states of the subjects change along with the stimuli \cite{9698041}. Cutting each trial into short segments and assigning the same label to them will induce noisy labels \cite{9091308}. To relieve noisy label issues and learn temporal contextual information of emotions, we can use a longer sliding window with short-shifting steps to split each trial, and the longer segments are further cut into sub-segments. Multi-head self-attention (MSA) in transformers \cite{NIPS2017_3f5ee243} can attentively emphasis the parts that are highly correlated to the overall emotional state of the longer EEG segment. Because emotion is continuous \cite{doi:10.1177/1754073915590618}, the underlying state is consistent in a short period. Hence, we propose a short-time aggregation (STA) layer after MSA to learn the long-short-time contextual information. 
Different from classification tasks, the label is temporally continuous in regression tasks. The model needs to regress the continuous changes of the emotional states for all the segments within a sequence. Although MSA can globally emphasis important parts in the sequence, recurrent neural networks (RNN) can further fuse the information of all the segments recurrently. This causal information fusion ability makes RNN more suitable for regression task. Hence, we propose to use an RNN-based token mixer in TCT for the regression tasks instead of MSA. To this end, we propose TCT-Clas and TCT-Regr as the token mixers in EmT for EEG emotion classification and regression tasks. 

\drop{The major contribution of this work can be summarised as: 
\begin{itemize}
\item We propose a novel Emotion Transformer (EmT) for generalized cross-subject EEG emotion classification and regression tasks.
\item We introduce a residual multi-view pyramid GCN (RMPG) module, designed to learn multi-view and multi-level graph embeddings using multiple learnable adjacency matrices. This approach incorporates neuroscience knowledge, recognizing that emotions are composed of various basic cognitive processes.
\item Two types of temporal contextual transformer (TCT) blocks with task-specific token mixers are proposed to capture temporal contextual information from EEG data, accounting for the differing learning objectives between classification and regression tasks.

\item Extensive experiments are conducted to evaluate and analyze the proposed EmT on four public datasets. Shanghai Jiao Tong University emotion EEG dataset (SEED) \cite{zheng2015investigating}, the TsingHua University emotional profile dataset (THU-EP) \cite{hu2022similar} and the finer-grained affective computing EEG dataset (FACED) \cite{Chen2023} are used for classification tasks while the multimodal database for affect recognition and implicit tagging dataset (MAHNOB-HCI) \cite{7112127} is utilized for regression tasks. The results demonstrate the superior of EmT over the compared baseline methods. 
\end{itemize}}

\section{Related Work}
\subsection{Graph Neural Networks}

Graph neural networks (GNN) are used for non-Euclidean graph-structure data. Spectral GNN is a category of GNN that often rely on expensive eigendecomposition of the graph laplacian, thus several methods use approximation approaches to perform spectral filtering. ChebyNet \cite{NIPS2016_04df4d43} uses Chebyshev polynomials to approximate the spectral filters. Cayley polynomials are utilized to compute spectral filters for targeted frequency bands, which are localized in space and scale linearly with input data for sparse graphs \cite{8521593}. The graph convolutional network (GCN) approximates spectral filtering with localized first-order aggregation \cite{kipf2017semisupervised}. EEG signals naturally have a graph structure. While some methods use GNN/GCN to extract spatial information from EEG signals, several approaches \cite{9091308,8320798,8815811} do not consider the interactions among multiple brain areas involved in high-level cognitive processes, relying instead on a single adjacency matrix. Additionally, most of these methods \cite{9091308,8320798,8815811,Song_Liu_Zheng_Zong_Cui_2020} neglect the temporal contextual information associated with emotional processes, using averaged features as node attributes. To effectively extract spatial relationships among EEG channels, we employ ChebyNet as the GCN layer in our model. Moreover, instead of solely using averaged features, we construct EEG signals as a sequence of spatial graphs to explicitly learn the temporal contextual information.

\subsection{Temporal Context Learning}
Emotion is a continuous cognitive process underlying which the temporal contextual information is embedded in the EEG signals. Methods that can learn the temporal dynamics of the sequence are often used as a temporal context extraction module. \cite{7112127} utilize a long short-term memory network (LSTM) to predict the temporally continuous emotion scores. Using a temporal convolutional network (TCN) shows improved emotion regression results in \cite{ZHANG2022108833}. However, both of them learn from flattened EEG feature vectors, which cannot effectively learn the spatial relations. TESANet \cite{9892920} uses 1-D CNN to extract spatial information in spectral filtered EEG after which LSTM and self-attention are utilized to extract temporal dynamics to predict the odor pleasantness from the EEG signal. Conformer \cite{9991178} combines CNN and transformer, achieving promising classification results for both emotion and motor imagery tasks. AMDET \cite{10261214} utilizes a transformer and attention mechanism on the spectral-spatial-temporal dimensions of EEG data for EEG emotion recognition.
Different from them, our model effectively learns from spatial topology information via parallel GCNs with learnable adjacency matrices. We also propose a short-time aggregation layer inspired by the prior knowledge that emotion is short-term stable and long-term varying. 

\subsection{EEG Emotion Recognition}
Emotion recognition using EEG data presents a formidable challenge, primarily due to the inherent variability across subjects and the subjectivity involved in perceiving emotions. The efficacy of prediction hinges on the development of a model capable of discerning crucial features that can distinguish between emotional classes. These extracted features typically encompass spectral, spatial, and temporal characteristics. Spectral features are typically derived through Fourier-based techniques, involving data filtering, and subsequent calculation of parameters like power spectral density (PSD) or differential entropy (DE). Alternatively, spectral filtering can be achieved by convolutions along the temporal dimension. Spatial features, on the other hand, are obtained through spatial convolution, often using convolutional neural networks (CNNs) and graph convolutional neural networks (GNNs). \cite{8320798} propose DGCNN, a GCN-based network with a learnable adjacency matrix, to learn dynamical spatial patterns from differential entropy (DE) features. Based on DGCNN, a broad learning system (BLS) is added in graph convolutional broad network (GCB-Net) \cite{8815811}, which improves the emotion classification results. \cite{9091308} proposed a regularized graph neural network (RGNN) with a neuroscience-inspired learnable adjacency matrix that is constrained to be symmetric and sparse to perform graph convolution. \cite{Song_Liu_Zheng_Zong_Cui_2020} proposed the instance-adaptive graph method to create graphic connections that are adapted to the given input. TSception \cite{9762054} utilizes multi-scaled temporal and spatial kernels to extract multiple frequency and spatial asymmetry patterns from EEG. Although they can learn the spatial and short-time temporal patterns, less attention is paid to the long-short-time contextual information underlying emotional cognitive processes.

\section{Method}
In this work, we propose a novel transformer, EmT, for both the generalized cross-subject EEG emotion classification and regression tasks. The network architecture is shown in Figure~\ref{fig:EmT}. EmT consists of four main parts: (1) temporal graph construction module (TGC), (2) RMPG, (3) TCT, and (4) task-specific output module (TSO). In TGC, EEG signals are transformed into temporal graph format that is a sequence of EEG feature graphs, as shown in Figure~\ref{fig:temporal_graph}. RMPG learns dynamical graph representations for each EEG feature graph within the series and the learned representations of each graph are fused into one token. TCT learns the temporal contextual information via specially designed token mixers. Finally, the TSO module will generate the desired output for classification and regression tasks accordingly. 

\subsection{EEG-temporal-graph Representations}
Temporal graphs are constructed to allow the neural network to learn spatial and temporal contextual information. Two steps are required to generate the temporal graphs from EEG, which are EEG segment/sub-segment segmentation and feature extraction, respectively. Figure~\ref{fig:temporal_graph} shows the construction process for one EEG segment.

Firstly, EEG signals are split into short segments that are further split into several shorter sub-segments using sliding windows. Given one trial of $c$-channel EEG signals, denoted by $\boldsymbol{X} \in \mathbb{R}^{c \times L}$, it is split into short segment $\Bar{\boldsymbol{X}} \in \mathbb{R}^{c \times l}$ using a sliding window of length $l$, with a hop step being $s$. Then another sliding window whose length and hop step are ${l}'$ and ${s}'$ is utilized to split each $\Bar{\boldsymbol{X}}$ into a series of sub-segment $\tilde{\boldsymbol{X}} \in \mathbb{R}^{c \times {l}'} $, where $L > l > {l}'$. In this paper, $l=20 sec=20*f_{s}, s=4 sec=4*f_{s}$ for all three datasets, where $f_{s}$ is the sampling rate of EEG signals. For SEED and THU-EP, $ {l}'=2 sec=2*f_{s}$ and $ {s}'=0.5 sec=0.5*f_{s}$. For FACED, $ {l}'=4 sec=4*f_{s}$ and $ {s}'=1 sec=1*f_{s}$. 

Relative power spectral density (rPSD) features are calculated for each sub-segment $\tilde{\boldsymbol{X}}$ to extract frequency information of short-period EEG signals. Specifically, the rPSD in delta (1-4 Hz), theta (4-8 Hz), alpha (8-12 Hz), low beta (12-16 Hz), beta (16-20 Hz), high beta (20-28 Hz), and gamma (30-45 Hz) seven bands are calculated using welch's method for each EEG channel to get a feature matrix $\boldsymbol{F} \in \mathbb{R}^{c \times f}$, where $f=7$, for each $\tilde{\boldsymbol{X}}$. Each channel is regarded as one node and the rPSDs are regarded as node attributes. Hence, we have a temporal graph representation $\boldsymbol{G}_{T} = \{\mathcal{G}^{i}\} \in \mathbb{R}^{seq \times c \times f}$ for one $\Bar{\boldsymbol{X}}$. 

\subsection{Residual Multi-view Pyramid GCN}
An RMPG is proposed to modulate the dynamical spatial relations among EEG channels underlying emotional processes. For each $\mathcal{G}^{i}$ in $\boldsymbol{G}_{T}$, RMPG learns one flattened embedding as one temporal token $\boldsymbol{s}^{i}$ for the subsequent transformer model. 

A base graph encoder, $\Phi_{g}(\cdot)$, is utilized to learn graph representations. $\Phi_{g}(\cdot)$ can be ChebyNet \cite{NIPS2016_04df4d43}, GCN \cite{kipf2017semisupervised}, GAT \cite{veličković2018graph} etc.. In this paper, we use ChebyNet as our base graph encoder:
\begin{equation}\label{eq:Cheby}
    \Phi_{g}(\boldsymbol{F}^{m}, \boldsymbol{A})= \sigma(\sum_{k=0}^{K-1}\boldsymbol{\theta}_{k}^{m}T_{k}(\hat{\boldsymbol{L}})\boldsymbol{F}^{m-1} - \boldsymbol{b}^{m}),   
\end{equation}
where $m=[1, 2, ...]$ is the number of GCN layers, $\boldsymbol{A} \in \mathbb{R}^{c \times c}$ is the adjacency matrix, $\sigma$ is the ReLU activation function, $\boldsymbol{\theta}$ is the learnable parameter, $T_{k}$ is the $k$-order Chebyshev polynomials that are defined as $T_{0}(x)=1$, $T_{1}(x)=x$, $T_{i}(x)=2T_{i-1}(x) - T_{i-2}(x)$, $\hat{\boldsymbol{L}} = 2\boldsymbol{L}/\lambda_{max} - \boldsymbol{I}$ with $\boldsymbol{L} = \boldsymbol{I} - \boldsymbol{D}^{-\frac{1}{2}}\boldsymbol{A}\boldsymbol{D}^{-\frac{1}{2}}$ is the re-scaling operation of $\boldsymbol{L}$, and $\boldsymbol{b}$ is the bias. We approximate $\lambda_{max} \approx 2$ as \cite{kipf2017semisupervised} to remove its high computational cost. Hence, we have $\hat{\boldsymbol{L}} = - \boldsymbol{D}^{-\frac{1}{2}}\boldsymbol{A}\boldsymbol{D}^{-\frac{1}{2}}$.

To modulate different brain region connections for multiple basic cognitive processes underlying emotions \cite{KOBER2008998}, we propose to use multiple different-layer GCNs, \{$\Phi_{g}^{0}(\cdot), \Phi_{g}^{1}(\cdot),...,\Phi_{g}^{i}(\cdot)$\} with learnable adjacency matrices \cite{8320798}, \{$\boldsymbol{A}^{0}, \boldsymbol{A}^{1},...,\boldsymbol{A}^{i}$\}. 
\drop{These adjacency matrices are randomly initialized with different values and can be dynamically adjusted using gradient backpropagation \cite{8320798}.} Each $\boldsymbol{A}^{i}$ can learn one view of graph connections that belongs to a certain basic cognitive process. For each $\Phi_{g}^{i}(\cdot)$, stacking different layers of GCNs can learn different degrees of node cluster similarity \cite{Li_Han_Wu_2018}. Intuitively, for localized connections, such as electrodes within a brain functional area, a deeper GCN can get a consistent representation among these nodes. However, for global connections that are among different brain functional areas, a shallow GCN can aggregate the information among these areas while not over smooth it. The deeper the $\Phi_{g}^{i}(\cdot)$ is, the taller the feature pyramid level its output has. A linear projection layer, $\textrm{LP}(\cdot)$, is added for each $\Phi_{g}^{i}(\cdot)$ to project the flattened graph representations into a hidden embedding denoted by $\boldsymbol{H}_{g}^{i} \in \mathbb{R}^{d_{g}}$. By stacking $\boldsymbol{H}_{g}^{i}$ from parallel different-layer GCNs, we can get multi-view pyramid graph embeddings as:
\begin{equation}\label{eq:graph_mv}
    \{\boldsymbol{H}_{g}^{i}\}=\{\textrm{LP}^{i}(\Gamma(\Phi_{g}^{i}(\boldsymbol{F}, \boldsymbol{A}^{i})))\},   
\end{equation}
where $\{\cdot\}$ is the stack operation and $\Gamma(\cdot)$ is the flatten operation.

A linear residual branch is added to additionally provide some information about the non-filtered graphs. The output serves as the base of the feature pyramid. The linear layer project the flattened $\mathcal{G}^{i}$ into a vector, $\boldsymbol{H}_{g-base}$, that has the same size as $\boldsymbol{H}_{g}^{i}$. A mean fusion is applied to combine different level graph information in the feature pyramid to form one token:
\begin{equation}\label{eq:graph_fuse}
      \boldsymbol{s}=\textrm{mean}(\{\boldsymbol{H}_{g-base},\boldsymbol{H}_{g}^{0}, ..., \boldsymbol{H}_{g}^{i}\}),
\end{equation}
where $\boldsymbol{s} \in \mathbb{R}^{d_{g}}$ is the token embedding for each $\mathcal{G}$ of $\boldsymbol{G}_{T}$, and $\boldsymbol{H}_{g-base}=\textrm{LP}(\Gamma(\mathcal{G}))$. To this end, $\boldsymbol{G}_{T}$ becomes a temporal token sequence denoted by $ \boldsymbol{S}_{T}=\{\boldsymbol{s}^{i}\} \in \mathbb{R}^{seq \times d_{g}}$.

\subsection{Temporal Contextual Transformer}
Two types of TCTs with different token mixers for classification and regression tasks are proposed to capture temporal contextual information of EEG underlying emotional processes. MetaFormer \cite{Yu_2022_CVPR} shows that MSA can be replaced by different token mixers in transformers. We propose two types of token mixers with which we have EmT-Clas and EmT-Regr for EEG emotion classification and regression separately. EmT refers to EmT-Clas unless otherwise stated. Given $\boldsymbol{S}_{T}=\{\boldsymbol{s}^{i}\} \in \mathbb{R}^{seq \times d_{g}}$ from RMPG, one block of TCT can be represented as:
\begin{equation}\label{eq:TCT_1}
      \boldsymbol{Z}^{m}= \textrm{TokenMixer}_{\textrm{clas/regr}}(\textrm{Norm}(\boldsymbol{Z}^{m-1})) + \boldsymbol{Z}^{m-1},
\end{equation}
\begin{equation}\label{eq:TCT_2}
      \boldsymbol{Z}^{m+1}= \textrm{MLP}(\textrm{Norm}(\boldsymbol{Z}^{m})) + \boldsymbol{Z}^{m},
\end{equation}
where $m=[1, 2, ...]$ is the number of layers in TCT blocks, $\boldsymbol{Z}^{0}=\boldsymbol{S}_{T}$, and $\textrm{MLP}$ has two linear layers with the ReLU activation in between. A dropout layer is added after each linear layer. 

\subsubsection{Token Mixers for Classification Tasks}
For classification tasks, MSA is utilized in $\textrm{TokenMixer}_{\textrm{clas}}(\cdot)$ to attentively emphasis the parts that are highly correlated to the overall emotional state of $\boldsymbol{S}_{T}$. The tokens in $\boldsymbol{S}_{T}$ are then linearly projected into multiple groups of key ($\boldsymbol{K}^{i}$), query ($\boldsymbol{Q}^{i}$), and value ($\boldsymbol{V}^{i}$) using multiple $\textrm{LP}(\cdot)$, parameterized by $\boldsymbol{W}^{i}_{qkv} \in \mathbb{R}^{d_{g} \times 3d_{head}}$:
\begin{equation}\label{eq:Emb_qkv}
      \{\boldsymbol{Q}^{i}, \boldsymbol{K}^{i}, \boldsymbol{V}^{i}\}=\textrm{LP}^{i}(\boldsymbol{S}_{T})=\boldsymbol{S}_{T}\boldsymbol{W}^{i}_{qkv}.
\end{equation}
The scaled dot-product is utilized as the attention operation along temporal tokens to capture long-time context:
\begin{equation}\label{eq:attention}
      \textrm{Attn}(\boldsymbol{Q}, \boldsymbol{K}, \boldsymbol{V}) =\textrm{Softmax}(\boldsymbol{Q}\boldsymbol{K}^{T}/\sqrt{d})\boldsymbol{V},
\end{equation}
where $d=d_{head}$ is a scaling factor. Because we need to apply the proposed STA on separate outputs from heads, we just stack the head outputs in the formula below:
\begin{equation}\label{eq:MSA}
      \textrm{MSA}(\boldsymbol{S}_{T}) = \{\textrm{Attn}(\textrm{LP}^{0}(\boldsymbol{S}_{T}), ..., \textrm{Attn}(\textrm{LP}^{n_{head}-1}(\boldsymbol{S}_{T})\}.
\end{equation}
Considering the fact that emotion is short-term continuous and long-term varying \cite{9698041}, we propose a short-time aggregation (STA) layer after MSA to learn the long-short-time contextual information. A dropout layer with a scaling factor $\alpha$, $\textrm{dp}(\alpha)$ is applied to control the forgetting rate of the temporal context. $\alpha$ is a hyper-parameter that will scale down the overall dropout rate in STA. To capture short-term consist patterns and smooth the effects of the changes in emotional states, CNN kernels denoted by $\boldsymbol{K}_{cnn}$ whose size and step are $(n_{anchor}, 1)$, $(1, 1)$ are utilized to aggregate $n_{anchor}$ temporal neighbors after MSA. Let $\boldsymbol{H}_{attn} \in \mathbb{R}^{n_{head} \times seq \times d_{head}}$ denotes one output of MSA. STA can be described as:
\begin{equation}\label{eq:STA}
      \textrm{STA}(\boldsymbol{H}_{attn}) = \textrm{Reshape}(\textrm{Conv2D}(\textrm{dp}(\boldsymbol{H}_{attn}), \boldsymbol{K}_{cnn}))\boldsymbol{W}_{sta},
\end{equation}
where $\textrm{Conv2D}(\cdot)$ is the 2D convolution operation with the input being $\textrm{dp}(\boldsymbol{H}_{attn})$ and the kernel being $\boldsymbol{K}_{cnn}$, $\textrm{Reshape}(\cdot)$ is the reshape operation $(n_{head}, seq, d_{head}) \rightarrow (seq, n_{head}*d_{head})$, and $\boldsymbol{W}_{sta} \in \mathbb{R}^{n_{head}*d_{head} \times d_{g}}$ is the projection weight matrix. The input and output channels of the $\textrm{Conv2D}(\cdot)$ all equal $n_{head}$. The same padding is utilized to avoid size changes before and after STA. Hence, $\textrm{TokenMixer}_{\textrm{clas}}$ can be described as:
\begin{equation}\label{eq:token_mixer_clas}
      \textrm{TokenMixer}_{\textrm{clas}}(\boldsymbol{S}_{T})=\textrm{STA}(\textrm{MSA}(\boldsymbol{S}_{T})).
\end{equation}

\subsubsection{Token Mixers for Regression Tasks} 
An RNN-based token mixer is utilized instead of MSA for $\textrm{TokenMixer}_{\textrm{regr}}(\cdot)$. MSA emphasises the parts that are highly correlated to the overall emotional state within a sequence via SA. This is helpful for the classification task that requires a single output of the sequence. However, the regression task needs the model to predict the continuous changes in the emotional states for all the segments within the sequence. Because RNN-based token mixer can fuse the information of all the segments recurrently, it is more suitable for the regression task. The tokens in $\boldsymbol{S}_{T} \in \mathbb{R}^{seq \times d_{g}}$ are projected into values ($\boldsymbol{V}$) using a projecting weight matrix, $\boldsymbol{W}_{v} \in \mathbb{R}^{d_{g} \times d_{head}}$. Because the RNN-based token mixer can be RNN, LSTM, GRU, etc., we use RNNs to denote the RNN family. A two-layer bi-directional GRU whose output length is $2*d_{head}$ is empirically selected as the token mixer for EmT-Regr in this paper. Hence, $\textrm{TokenMixer}_{\textrm{regr}}$ can be described by:
\begin{equation}\label{eq:token_mixer_regr}
      \textrm{TokenMixer}_{\textrm{regr}}(\boldsymbol{S}_{T})=\textrm{RNNs}(\textrm{LP}(\boldsymbol{S}_{T})),
\end{equation}
where $\textrm{LP}(\boldsymbol{S}_{T})=\boldsymbol{S}_{T}\boldsymbol{W}_{v}$ is a linear layer. 
\subsection{Task-specific Output Module}
MLP heads are utilized to generate the desired output for classification and regression tasks. Let $\boldsymbol{S}_{clas} \in \mathbb{R}^{seq \times d_{head}}$ and $\boldsymbol{S}_{regr}\in \mathbb{R}^{seq \times 2*d_{head}}$ denote the learned embedding sequences for classification and regression tasks. The difference is that a mean fusion is applied to $\boldsymbol{S}_{clas}$ to combine the information of all the segments. Hence, the final classification output, $\hat{\boldsymbol{Y}}_{clas} \in \mathbb{R}^{n_{class}}$ is calculated by:
\begin{equation}\label{eq:to_out_clas}
      \hat{\boldsymbol{Y}}_{clas} = \textrm{mean}(\boldsymbol{S}_{clas})\boldsymbol{W}_{clas} + \boldsymbol{b}_{clas}
\end{equation}
where $\boldsymbol{W}_{clas} \in \mathbb{R}^{d_{head} \times n_{class}}$, and $\boldsymbol{b}_{clas} \in \mathbb{R}^{n_{class}}$ are the weights and bias, respectively. And the final regression output, $\hat{\boldsymbol{Y}}_{regr} \in \mathbb{R}^{seq}$ is calculated by:
\begin{equation}\label{eq:to_out_regr}
      \hat{\boldsymbol{Y}}_{regr} = \boldsymbol{S}_{regr}\boldsymbol{W}_{regr} + b_{regr}
\end{equation}
where $\boldsymbol{W}_{regr} \in \mathbb{R}^{2*d_{head}\times 1}$, and $b_{regr} \in \mathbb{R}^{1}$ are the weights and bias, respectively.

\begin{table}[tp] \centering\arraybackslash
\caption{Details of EmT variants.}
\label{tab:model_variants}
\begin{adjustbox}{center}
\begin{tabular}{lwc{3em}wc{3em}wc{3em}wc{3em}wc{3em}wc{3em}}
\toprule
               Model & GCN layers$^{\star}$ & $d_{g}$ & TCT layers& $n_{head}$ &$d_{head}$& $K^{\dagger}$\\ \midrule
		      EmT-S & 1, 2 & 32 & \textbf{2} &  16& 32& 3/4\\
                EmT-B & 1, 2 & 32 & \textbf{4} &  16& 32& 3/4\\
                EmT-D & 1, 2 & 32 & \textbf{8} &  16& 32& 3/4\\
               \bottomrule
			\end{tabular}
		\end{adjustbox}
		\begin{tablenotes}
			\item[1] $\star$: the layers of two parallel GCNs; $\dagger$: $K=3$ for less-channel EEG (e.g. 32 in THU-EP and MAHNOB-HCI) and $K=4$ for more-channel EEG (e.g. 62 in SEED).
		\end{tablenotes}
	\end{table}

\begin{table*}[htp] \centering\arraybackslash
\caption{Generalized emotion classification results of different methods on the SEED, THU-EP and FACED datasets. The best results are highlighted in bold and the next best are marked using underlines.}
\label{tab:classification_results}
\begin{adjustbox}{center}
\begin{tabular}{lwc{3em}wc{3em}wc{3em}wc{3em}wc{3em}wc{3em}wc{3em}wc{3em}wc{3em}wc{3em}wc{3em}wc{3em}}
\toprule
\multirow{2}{*}{Method} & \multicolumn{4}{c}{SEED} & \multicolumn{4}{c}{THU-EP}  & \multicolumn{4}{c}{FACED}              \\ \cmidrule(lr){2-5} \cmidrule(lr){6-9} \cmidrule(lr){10-13}
                & ACC & std & F1& std& ACC& std & F1 & std & ACC& std & F1 & std\\ \midrule
		     DGCNN & 0.724 &0.145 & 0.619 & 0.311 & 
              0.567 & 0.033 &  0.647 & 0.052 &0.562&0.045&0.697&0.043\\
              GCB-Net & 0.684 & 0.172 & 0.517 &0.357 & 0.554 &0.043 &  0.620 &0.088 &0.565&0.052 &0.685 &0.053\\
              RGNN & \underline{0.790} &0.148 & \underline{0.802} &0.133
              & 0.572 &0.030 &  0.695 &0.054 &0.587&0.050&0.722&0.721\\
              TSception & 0.662 &0.181 & 0.621 &0.283 & 0.591 &0.059 &  \underline{0.707} &0.060 &\textbf{0.619}&0.088&0.702& 0.237\\
              TCN & 0.765 &0.140 & 0.737 &0.219 & 0.577 &0.031 &  0.677 &0.031 &0.552 &0.035&0.673&0.035\\
              LSTM & 0.733 & 0.158 & 0.670 &0.274 & 0.558 & 0.035 &  0.626 &0.062 & 0.568&0.063&0.700&0.064\\
              TESANet & 0.658 &0.115 & 0.606 & 0.240 & 0.578 &0.037& 0.673 &0.072 &0.593&0.054&0.723&0.056\\ 
              Conformer &0.612  &0.127 & 0.529 & 0.221 & \textbf{0.601}  &0.038 & 0.691 & 0.042&0.590&0.035&0.720&0.035\\
              AMDET &0.721  &0.168 &0.649 &0.306 &0.586 &0.044 &0.677 &0.098 &0.591& 0.043& \underline{0.726}& 0.043\\
              \midrule
              EmT-S (ours) & 0.780 &0.117 & 0.759 &0.159 & 0.570 &0.032 & 0.662 &0.050 &0.589&0.029&0.723&0.030\\ 
              EmT-B (ours) & 0.788 &0.117 & 0.793 &0.120 & \underline{0.595} &0.047 & \textbf{0.724} & 0.044 &\underline{0.608}& 0.065&\textbf{ 0.740}& 0.058\\ 
              EmT-D (ours) & \textbf{0.802} &0.115 & \textbf{0.821} &0.093 & 0.583 &0.049 & 0.671 &0.102 &0.579&0.068&0.709&0.062\\ \bottomrule
			\end{tabular}
		\end{adjustbox}
	\end{table*}
\section{Experiment}
\subsection{Datasets}
For the classification task, we evaluate the performance of EmT with three emotion EEG datasets, which are SEED \cite{zheng2015investigating}, THU-EP \cite{hu2022similar} and FACED \cite{Chen2023}. For the regression task, we use a subset of the MAHNOB-HCI dataset \cite{7112127}.

The SEED dataset encompasses data from 15 native Chinese subjects, with the objective of eliciting negative, positive, and neutral emotions using 15 Chinese film clips. Each film clip has a duration of approximately 4 minutes. Subsequently, the participants are tasked with providing self-evaluations regarding their emotional responses, considering dimensions such as valence and arousal after viewing these film clips. To record brain activity, EEG signals are acquired using a 62-channel electrode setup arranged in the 10-20 system, at a high sampling rate of 1000 Hz. The data undergo preprocessing, which includes applying a bandpass filter with a frequency range of 0.3 to 50 Hz.

The THU-EP dataset comprises data from 80 subjects, involving the use of 28 video clips as stimuli to elicit negative, positive, and neutral emotions. Each video clip has an average duration of approximately 67 seconds. These video clips are associated with a range of emotion items, including anger, disgust, fear, sadness, amusement, joy, inspiration, tenderness, arousal, valence, familiarity, and liking. After viewing each video clip, subjects provided self-report emotional scores for these emotion items. The EEG data was recorded using a 32-channel EEG system at a sampling rate of 250 Hz. The collected data underwent preprocessing, which included applying a bandpass filter with a frequency range of 0.05 to 47 Hz. Additionally, independent component analysis (ICA) was employed to effectively remove artifacts from the EEG data.

The FACED dataset is an extended version of the THU-EP dataset, comprising data from 123 subjects. The experimental protocol remains the same as THU-EP, with the addition of 43 subjects, making it a relatively larger dataset for studying emotions using EEG. The official pre-processed data is used in this study, with pre-processing steps identical to those in THU-EP.

MAHNOB-HCI is a comprehensive multi-modal dataset designed for the investigation of human emotional responses and the implicit tagging of emotions. This dataset involves the participation of 30 subjects in data collection experiments. In these experiments, each subject views 20 film clips while a variety of data streams are recorded in synchronization. For the specific task of emotion recognition, a subset of the MAHNOB-HCI database, as referenced in [1], is employed. This subset encompasses 24 participants and 239 trials, with continuous valence labels provided by multiple experts. The final labels used for analysis are derived by averaging the annotations from these experts. The EEG signals in the dataset are collected using 32 electrodes, and they are sampled at a rate of 256 Hz. Notably, the annotations in the dataset are characterized by a resolution of 4 Hz.

\subsection{Baselines}

We demonstrate the performance of EmT by comparing the following baseline methods:

\subsubsection{DGCNN (graph-based)}
DGCNN \cite{8320798} dynamically learns the relationships between different EEG channels through neural network training, represented by a learnable adjacency matrix. This dynamic learning enhances the extraction of more discriminative EEG features, ultimately improving EEG emotion recognition.
\subsubsection{GCB-Net (graph-based)}
Based on DGCNN, graph convolutional broad network (GCN-Net) \cite{8815811} integrates a broad learning system (BLS) to the nerual network. GCB-Net employs a graph convolutional layer to extract features from graph-structured input and then stacking multiple conventional convolutional layers to derive relatively abstract features. Subsequently, the final concatenation phase adopts a broad concept, preserving the outputs from all hierarchical layers, which facilitates the model in exploring features across a wide spectrum.

\subsubsection{RGNN (graph-based)}
RGNN \cite{9762054} utilizes the biological topology of different brain regions to capture both local and global relationships among EEG channels. It models the inter-channel relations in EEG signals through an adjacency matrix within a graph neural network, where the connections and sparsity of the adjacency matrix are inspired by neuroscience theories regarding human brain organization. 

\subsubsection{TSception (CNN-based)}
TSception \cite{9762054} is a multi-scale convolutional neural network designed for generalized emotion recognition from EEG data. TSception incorporates dynamic temporal, asymmetric spatial, and high-level fusion layers, working together to extract discriminative temporal dynamics and spatial asymmetry from both the time and channel dimensions.
\subsubsection{LSTM (temporal-learning)}
LSTM based neural networks \cite{7112127} are utilized to predict the continuously annotated emotional labels. LSTM is capable to learn the long-term temporal dependencies among the EEG segments. 
\subsubsection{TCN (temporal-learning)}
Zhang et al., use TCN \cite{ZHANG2022108833} to learn the temporal information for the continuous EEG emotion regression tasks. The results indicate TCN has better regression performances than LSTM. 
\subsubsection{TESANet (temporal-learning)}
TESANet \cite{9892920} has been developed to discern the relations between time segments within EEG data, enabling the prediction of pleasant and unpleasant emotional states. This network architecture comprises a filter-bank layer, a spatial convolution layer, a temporal segmentation layer responsible for partitioning the data into overlapping time windows, a LSTM layer for encoding these temporal segments, and a self-attention layer.

\subsubsection{Conformer (temporal-learning)}
The Conformer \cite{9991178} combines a CNN encoder and a transformer to capture both the short- and long-term temporal dynamics encoded in EEG signals, achieving promising results for emotion and motor imagery classification tasks.

\subsubsection{AMDET (temporal-learning)}
AMDET \cite{10261214} learns from 3D temporal-spectral-spatial representations of EEG signals. It utilizes a spectral-spatial transformer encoder layer to extract meaningful features from the EEG signals and employs a temporal attention layer to highlight crucial time frames, considering the complementary nature of the spectral, spatial, and temporal features of the EEG data.
\subsection{Experiment Settings}
We conduct generalized subject-independent settings in this study where test data information is never used during the training stage. For SEED, we use a leave-one-subject-out (LOSO) setting. For each step in LOSO, one subject's data is selected as test data. 80\% of the training data is used as training data, and the rest 20\% is used as validation data. For the THU-EP and FACED datasets, we adopt a leave-n-subject-out setting \cite{9748967}, where $n_{\textrm{THU-EP}}=8$, and $n_{\textrm{FACED}}=12$. The last three subjects are added in the 10-th fold in FACED. 10\% of the training data is used as the validation data. We perform binary classification on the positive and negative emotions on SEED, THU-EP, and FACED as \cite{9321519}. We process the valence score of THU-EP and FACED into a binary class of high and low valence using a threshold of 3.0. For the regression task, we follow the same data pre-processing and experiment settings used in \cite{ZHANG2022108833}. A LOSO is conducted for the regression task which is identical to the one for SEED.

\subsection{Model Variants}
We propose three types of EmT variants, namely EmT-shallow (EmT-S), EmT-base (EmT-B), and EmT-deep (EmT-D). The configurations are shown in Table \ref{tab:model_variants}. As the names show, the differences among these variants are the depth of the TCT blocks. For the $K$ of Chebyshev polynomials is decided by the number of EEG channels (nodes in graphs). This is because $K$ decides how many hops from the central vertex the GCN can aggregate. If the number of EEG channels is relatively less, e.g., 32 in THU-EP and MAHNOB-HCI, $K=3$. $K$ will be a larger value, $K=4$ when the EEG signals have more channels such as 62 in SEED. EmT-S is used as the EmT-Regr because it gives the best results on the validation (development) set. 

\subsection{Evaluation Metrics}
The evaluation metrics for emotion classification are the same as those in \cite{9762054}: Accuracy (ACC) and F1 scores. They can be calculated by

\begin{equation}\label{eq:acc}
     \textrm{Accuracy} = \frac{\textrm{TP}+\textrm{TN}}{\textrm{TP}+\textrm{FP}+\textrm{TN}+\textrm{FN}}
 \end{equation}

 \begin{equation}\label{eq:f1}
     \textrm{F1} = 2\times \frac{\textrm{Precision}\times \textrm{Recall}}{\textrm{Precision}+\textrm{Recall}}=\frac{\textrm{TP}}{\textrm{TP}+\frac{1}{2}(\textrm{FP}+\textrm{FN})}
 \end{equation}
where $\textrm{TP}$ denotes true positives, $\textrm{TN}$ denotes true negatives, $\textrm{FP}$ denotes false positives, and $\textrm{FN}$ denotes false negatives.

The evaluation metrics for emotion regression are the same as those in \cite{ZHANG2022108833}: root mean square error (RMSE), Pearson’s correlation coefficient (PCC), and concordance correlation coefficient (CCC). Given the prediction $\hat{\textbf{y}}$, and the continuous label $\textbf{y}$, RMSE, PCC, and CCC can be calculated by
\begin{equation}
    \textrm{RMSE} = \left\| \frac{\hat{\textbf{y}}-\textbf{y}}{N} \right\|^{2} = \sqrt{\frac{\sum_{N-1}^{i=0}(\hat{y}_{i}-y_{i})^{2}}{N}},
\end{equation}

\begin{equation}
    \textrm{PCC} = \frac{\sigma_{\hat{\textbf{y}}\textbf{y}}}{\sigma_{\hat{\textbf{y}}}\sigma_{\textbf{y}}}=\frac{\sum_{N-1}^{i=0}(\hat{y}_{i}-\mu_{\hat{\textbf{y}}})(y_{i}-\mu_{\textbf{y}})}{\sqrt{\sum_{N-1}^{i=0}(\hat{y}_{i}-\mu_{\hat{\textbf{y}}})^{2}}\sqrt{\sum_{N-1}^{i=0}(y_{i}-\mu_{\textbf{y}})^{2}}},
\end{equation}

\begin{equation}
    \textrm{CCC} = \frac{2\sigma_{\hat{\textbf{y}}\textbf{y}}}{\sigma_{\hat{\textbf{y}}}^{2} + \sigma_{\textbf{y}}^{2} + (\mu_{\hat{\textbf{y}}} - \mu_{\textbf{y}})},
\end{equation}
where $N$ denotes the number of elements in the prediction/label vector, $\sigma_{\hat{\textbf{y}}\textbf{y}}$ denotes the covariance, $\sigma_{\hat{\textbf{y}}}$ and $\sigma_{\textbf{y}}$ are the variances, and $\mu_{\hat{\textbf{y}}}$ and $\mu_{\textbf{y}}$ are the means.

\subsection{Implementation Details}
The model configurations can be found in Table \ref{tab:model_variants}. We first introduce the training parameters for classification tasks. The cross-entropy loss is utilized to guide the training. We use an AdamW optimizer with an initial learning rate of 3e-4. The label smoothing with a smoothing rate of 0.1 and a dropout rate of 0.25 are applied to avoid over-fitting. The batch size is 64 for all datasets. The training epochs are 10, 30, and 30 for SEED, THU-EP, and FACED. And the model with the best validation accuracy is used to evaluate the test data. $\alpha$ in STA is empirically selected as 0.25 for SEED, 0.1 for THU-EP and 0.4 for FACED. For regression tasks, we use a CCC loss, $\mathcal{L}_{CCC}(\hat{\textbf{y}},\textbf{y}) = 1-\frac{2\sigma_{\hat{\textbf{y}}\textbf{y}}}{\sigma_{\hat{\textbf{y}}}^{2} + \sigma_{\textbf{y}}^{2} + (\mu_{\hat{\textbf{y}}} - \mu_{\textbf{y}})}$, where $\sigma_{\hat{\textbf{y}}\textbf{y}}$ is the covariance, $\sigma_{\hat{\textbf{y}}}$ and $\sigma_{\textbf{y}}$ are the variances, and $\mu_{\hat{\textbf{y}}}$ and $\mu_{\textbf{y}}$ are the means, an Adam optimizer with an initial learning rate of 5e-5 and a weight decay of 1e-3, a batch size of 2, and a window length of 96 with a hop step of 32. We train the network for 30 epochs and use the model of the last epoch to evaluate the test data.

\begin{table}[t] \centering\arraybackslash
\caption{Emotion regression results on MAHNOB-HCI.} 
\label{tab:result_regression}
\begin{adjustbox}{center}
\begin{tabular}{lwc{4.2em}wc{4.2em}wc{4.2em}}
\toprule
 Method& RMSE $\downarrow$& PCC $\uparrow$ & CCC $\uparrow$ \\
\midrule 
LSTM \cite{7112127} & 0.081 & 0.427 & 0.306 \\ 
TCN \cite{ZHANG2022108833} & 0.066 & 0.474 & 0.377 \\
EmT-Regr (MSA)$^{\star}$ & 0.075 & 0.393 & 0.312 \\
EmT-Regr (LP+RNN)$^{\star}$ & 0.069 & 0.470 & 0.381\\
EmT-Regr (LP+LSTM)$^{\star}$ & \textbf{0.063} & 0.483 & 0.390 \\
EmT-Regr (LP+GRU)$^{\star}$ & 0.068 & \textbf{0.490} & \textbf{0.396}\\
\bottomrule
\end{tabular}
\end{adjustbox}
\begin{tablenotes}
      \small
      \item $\downarrow$: the lower the better; $\uparrow$: the higher the better.
      \item $^{\star}$:EmT-Regr (token mixer type)
    \end{tablenotes}
\end{table}

\begin{table}[t] \centering\arraybackslash
\caption{Generalized emotion classification results of ablation studies on the SEED and THU-EP datasets.}
\label{tab:ablation}
\begin{adjustbox}{center}
\begin{tabular}{lwc{4em}wc{4em}wc{4em}wc{4em}}
\toprule
\multirow{2}{*}{Method} & \multicolumn{2}{c}{SEED} & \multicolumn{2}{c}{THU-EP}                 \\ \cmidrule(lr){2-3} \cmidrule(lr){4-5}
                & ACC & F1 & ACC& F1 \\ \midrule
		       w/o RMPG & 0.775 & 0.793 & 0.577 &  0.663\\
                 w Single GCN & 0.773 & 0.760 & 0.551 &  0.642\\
                 w/o TCT & 0.777 & 0.749 & 0.592 &  0.698\\
                 w/o STA & 0.784 & 0.795 & 0.582 &  0.690\\ 
                 EmT (ours) & \textbf{0.802} & \textbf{0.821} & \textbf{0.595} &  \textbf{0.724}\\
               \bottomrule
			\end{tabular}
		\end{adjustbox}
	\end{table}

\section{Results and Analyses}
\subsection{Emotion Classification}

The experimental results are shown in Table \ref{tab:classification_results}. We evaluate the methods using accuracy and F1 score. On the SEED dataset, EmT-D achieves the highest accuracy (0.802) and the highest F1 score (0.821), indicating its effectiveness in emotion classification. EmT-B also performs well with an accuracy of 0.788 and an F1 score of 0.793, while EmT-S shows a slightly lower accuracy of 0.780 but a strong F1 score of 0.759. RGNN achieves the second-best performance on the SEED dataset with an accuracy of 0.790 and an F1 score of 0.802. Notably, the methods that use features as input generally perform better than those using EEG as input. Additionally, learning from the temporal sequence of features consistently achieves better performance than learning from the features directly, except in the case of RGNN. This indicates the effectiveness of learning the temporal contextual information. Furthermore, compared to TCN, LSTM, TESANet, and AMDET, EmT can learn the spatial information better with the help of the GCN-based modules.

The observations on the THU-EP and FACED datasets are different. For the THU-EP dataset, EmT-B achieves the best F1 score of 0.724, while Conformer achieves the best accuracy (0.601). On the FACED dataset, EmT-B leads with the second-best accuracy (0.608) and the highest F1 score (0.740). AMDET achieves the second-best F1 score of 0.726, while the best accuracy is achieved by TSception (0.619). As the classes are imbalanced in both the THU-EP and FACED datasets, F1 scores are more important than accuracy. EmT-B achieves the highest F1 scores on both THU-EP and FACED, demonstrating its effectiveness in emotion classification. Different from the observations on the SEED dataset, the baselines using EEG as input achieve better performance than those using features as input. This might be because there are more subjects in THU-EP and FACED, and using EEG directly can provide more information. 

\drop{For the SEED dataset (62 channels, 15 subjects), there is a positive correlation between the number of transformer layers and performance, with EmT-D (8 layers) performing best due to the enriched spatial information from the higher number of channels. However, for the THU-EP and FACED datasets (32 channels, more subjects), EmT-B (4 layers) achieves better results, likely due to fewer channels and higher subject variability, where deeper models like EmT-D tend to overfit. Overall, EmT-B strikes a better balance between temporal and spatial learning in datasets with fewer channels and greater inter-subject variability.}

\subsection{Emotion Regression}

The regression results are shown in Table \ref{tab:result_regression}. According to the results, EmT-Regr (LP+LSTM) achieved the lowest RMSE (0.063) among all compared approaches, while EmT-Regr (LP+GRU) achieved the best PCC (0.490) and CCC (0.396), indicating the effectiveness of the proposed method. However, when using MSA as the token mixer in EmT, the performance reduced dramatically, falling below all compared baselines. The difference between MSA and RNNs or TCN is that RNNs or TCN can fuse the information globally or locally, while MSA focuses on learning the global temporal relations and emphasizing certain parts of the sequence. Hence, the results indicate that fusing information from all segments is crucial for regression tasks.

\subsection{Ablation}
To explore the contributions of the RMPG, TCT, and STA modules, we conducted an ablation analysis by systematically removing each layer and observing the subsequent effects on classification performance. The results of this analysis are detailed in Table \ref{tab:ablation}. Among all the ablation experiments, using only a single GCN had the most detrimental impact, leading to a 2.9\% and 6.1\% decrease in ACC and F1 score on the SEED dataset, and a 4.4\% and 8.2\% decrease on the THU-EP dataset. When EmT was tested without the RMPG module, it exhibited the second-largest reduction in accuracy on both datasets, with decreases of 2.7\% and 1.8\%, respectively. Similarly, the removal of TCT and STA modules also resulted in noticeable reductions in classification accuracy and F1 scores. These findings highlight that all the proposed modules work synergistically to enhance the predictive capabilities of EmT. Furthermore, the ablation results underscore the significant contribution of the RMPG module, attributable to its ability to modulate dynamic spatial relations among EEG channels, which is crucial for effectively capturing emotional activity.

\subsection{Effect of EEG Features}
\begin{figure}[t]
\centering
    \subfigure[Classification]{
    \includegraphics[width=0.46\linewidth]{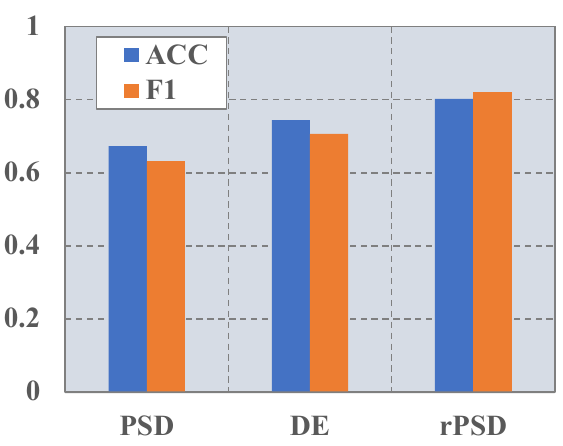}
    }
    \subfigure[Regression]{
    \includegraphics[width=0.46\linewidth]{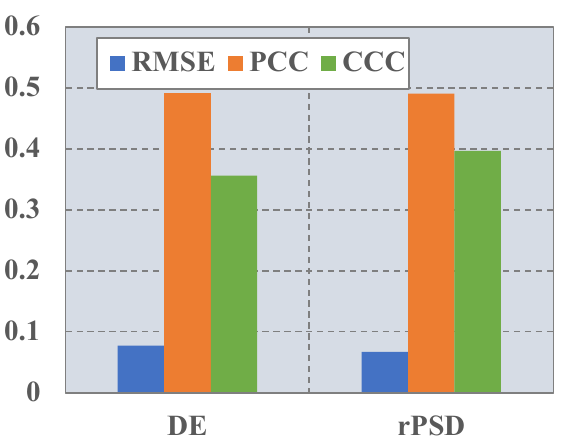}
    }
\caption{Effect of feature types on emotion classification and regression performances of EmT using SEED and MAHNOB-HCI. Using rPSD gives the overall best performances. We don't add PSD results for regression tasks in (b) because the model cannot converge.}
\label{fig:effect_feature}
\end{figure}

Fig.~\ref{fig:effect_feature} illustrates the impact of different feature types on the SEED dataset for classification tasks and the MAHNOB-HCI dataset for regression tasks. We compared three types of features: PSD, DE, and rPSD, evaluating their effects on accuracy and F1 score for classification tasks, as well as on MRSE, CCC, and PCC for regression tasks. For classification tasks, rPSD outperformed the other two feature types, providing the highest accuracy and F1 score. Specifically, rPSD achieved 5.9\% and 11.5\% higher accuracy and F1 score, respectively, compared to EmT using DE. Furthermore, rPSD showed a 13.0\% and 18.9\% improvement in accuracy and F1 score, respectively, over PSD. These results suggest that rPSD is a superior spectral feature for EEG emotion classification tasks.
For regression tasks, using rPSD resulted in better MRSE and CCC compared to using DE. However, the difference in PCC between rPSD and DE was minimal, with rPSD achieving a PCC of 0.491 and DE achieving a PCC of 0.490. Notably, when PSD was used as a feature, the model failed to converge, so we excluded these results from further analysis.
Based on these findings, rPSD is demonstrated to be a more effective feature than DE, and both rPSD and DE are superior to PSD for EEG emotion recognition tasks.

\begin{figure}[t]
\centering
    \subfigure[]{
    \includegraphics[width=0.465\linewidth]{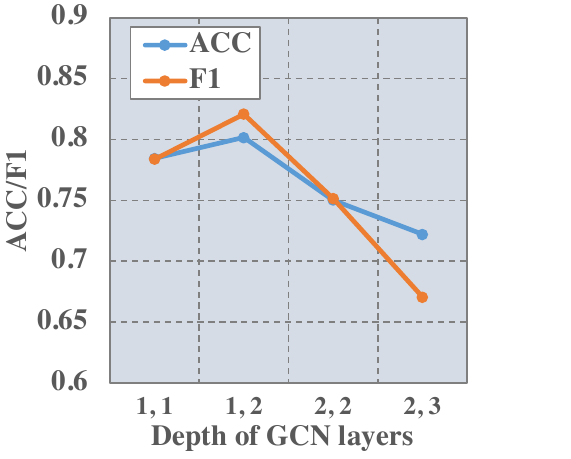}
    }
    \subfigure[]{
    \includegraphics[width=0.465\linewidth]{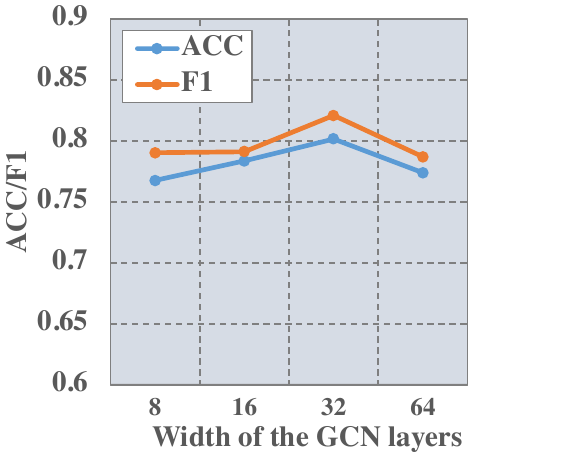}
    }
\caption{\drop{Effect of the depth (a) and width (b) of GCNs in RMPG on classification performance using SEED. For the depth analysis, 1,1 indicates that both GCN branches in RMPG have 1 layer each. The width is the hidden size of GCN layer.}}
\label{fig:effect_of_depth_width_GCN}
\end{figure}

\drop{\subsection{Effect of The Depth and Width of GCNs in RMPG}
We varied the number of GCN layers (depth) and the hidden size of the GCN layers (width) to evaluate their effects on classification performance. EmT-D and the SEED dataset were used for these experiments as they provide better overall performance compared to other configurations. The results are shown in Fig.~\ref{fig:effect_of_depth_width_GCN}. Since the RMPG module has two GCN branches, we tested four different configurations of GCN layers: [1, 1], [1, 2], [2, 2], and [2, 3]. These numbers represent the number of GCN layers (depth) for each branch in RMPG. As shown in Fig.~\ref{fig:effect_of_depth_width_GCN} (a), the [1, 2] configuration yields the best performance. Compared to [1, 1] and [2, 2], the results highlight the effectiveness of the pyramid design. Additionally, increasing the number of GCN layers leads to a significant drop in classification performance, consistent with the over-smoothing issue in deeper GCNs \cite{10540641}. In contrast, the effects of width are relatively smaller, as shown in Fig.~\ref{fig:effect_of_depth_width_GCN} (b). There is a positive correlation between width and performance when increasing the width from 8 to 32. However, the performance declines when the width is increased further, likely due to over-fitting caused by the larger model size.}

\subsection{Effect of The Number of TCT Blocks}
\begin{figure}[t]
\centering
    \subfigure[Classification]{
    \includegraphics[width=0.465\linewidth]{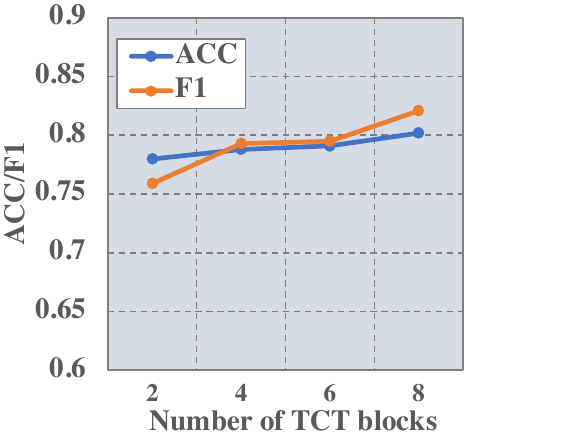}
    }
    \subfigure[Regression]{
    \includegraphics[width=0.465\linewidth]{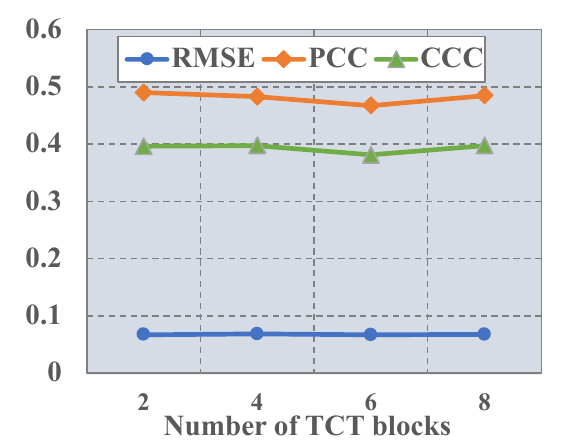}
    }
\caption{Effect of the number of TCT blocks on emotion classification and regression performances of EmT using SEED and MAHNOB-HCI.}
\label{fig:effect_of_TCT_layer}
\end{figure}

We vary the number of TCT blocks from 2 to 8 and monitor their effect on the results of the two tasks. The results are shown in Fig.~\ref{fig:effect_of_TCT_layer}. For classification, shown in Fig.~\ref{fig:effect_of_TCT_layer} (a), as the number of TCT blocks increases from 2 to 8, there is a noticeable improvement in both ACC and F1 scores. Specifically, accuracy increases from 0.780 to 0.802, while the F1 score shows a more substantial rise from 0.759 to 0.821, with the most significant improvement occurring between 6 and 8 TCT blocks. This indicates that adding more TCT blocks enhances the model's ability to capture temporal contextual information, thereby improving classification performance. Conversely, for regression tasks, shown in Fig.~\ref{fig:effect_of_TCT_layer} (b), the number of TCT blocks has little to no effect on the performance metrics. The RMSE remains stable around 0.06, and both the PCC and CCC show minimal variation, hovering around 0.48 and 0.39, respectively. This suggests that while increasing TCT blocks benefits classification by improving the capture of temporal contextual information, it does not significantly impact regression performance.

\begin{figure}[t]
    \centering
        \subfigure[SEED (62 Channels)]{
    \includegraphics[width=0.46\linewidth]{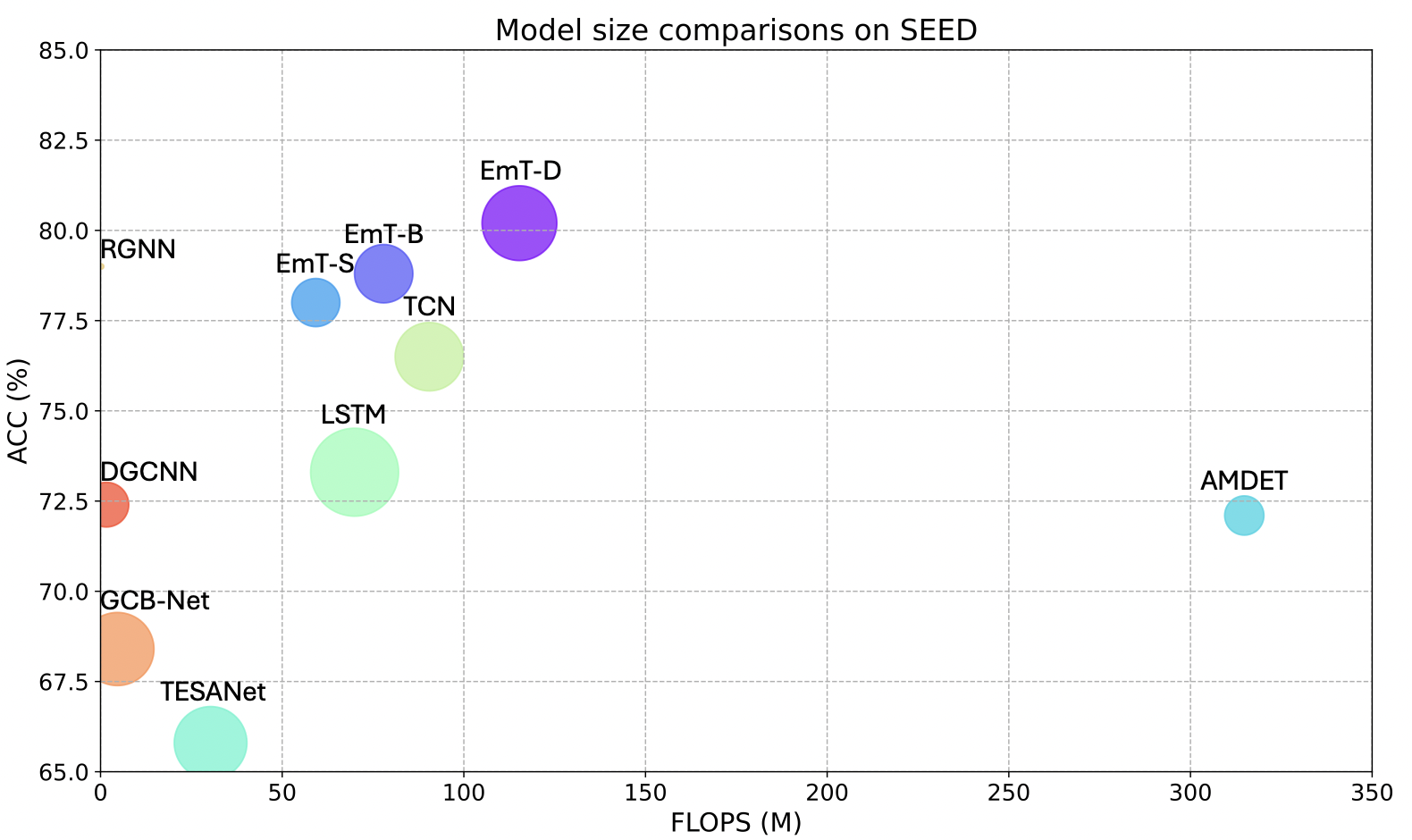}
    }
    \subfigure[FACED (32 Channels)]{
    \includegraphics[width=0.46\linewidth]{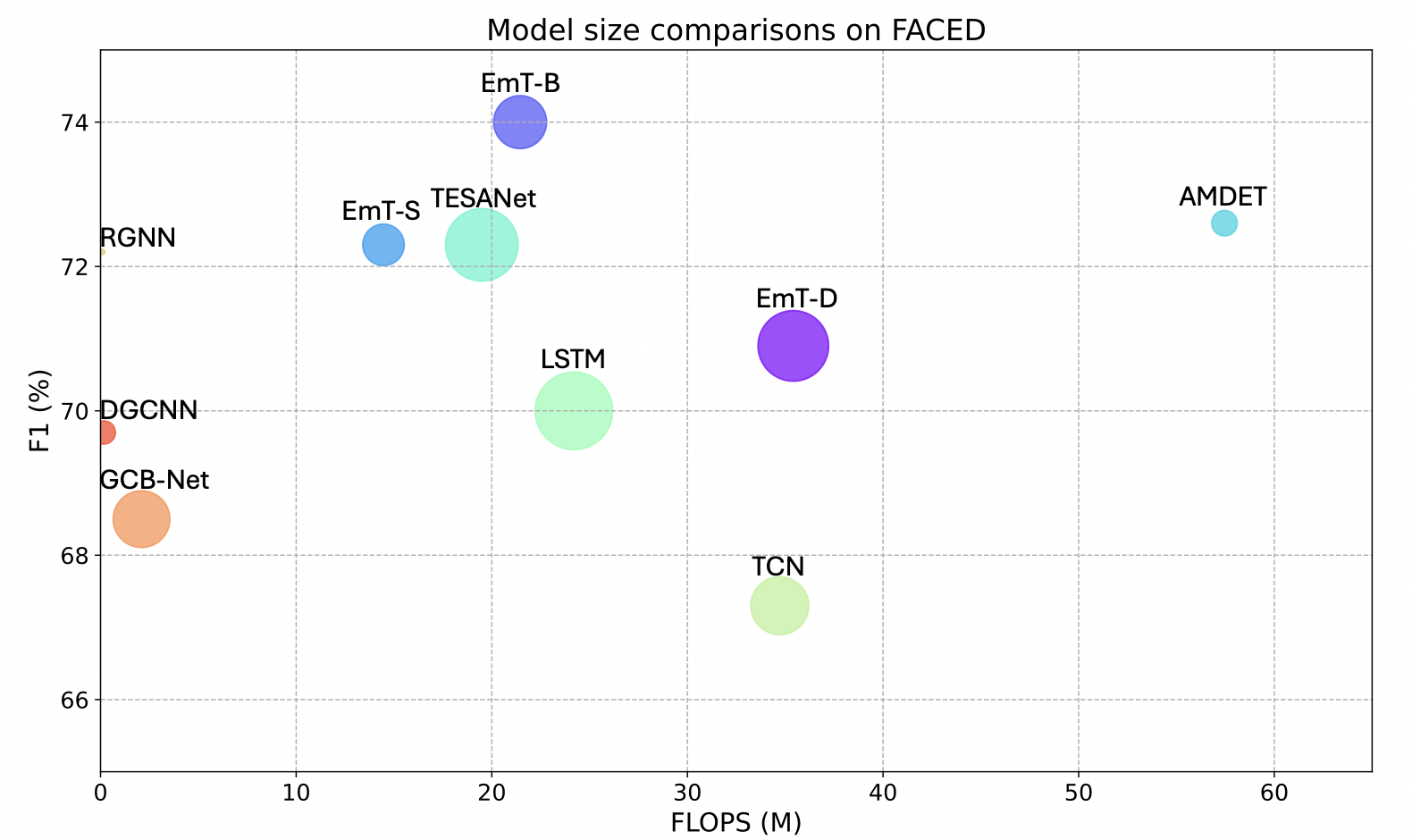}
    }
    \caption{\drop{Comparison of model performance, parameters, and FLOPS (M). The model size is represented by the area of the circle. TSception and Conformer are excluded due to much higher FLOPS than others. Our EmT-B achieves a promising balance between performance and computational complexity.}}
    \label{fig:flops}
\end{figure}

\drop{\subsection{Analysis of The Computational Complexity}
We conducted an analysis of the computational complexity of various models using floating-point operations (FLOPs) and the number of parameters as evaluation metrics. For the SEED dataset, ACC is used since the classes are balanced, while the F1 score is used for the FACED dataset due to its imbalanced classes. Since THU-EP and FACED share the same data format, we evaluated results only on FACED, which is larger than THU-EP. The results are shown in Fig~\ref{fig:flops}. TSception and Conformer are excluded from the comparison due to their significantly higher FLOPs than the other models. EmT-D achieves the highest ACC on SEED with relatively comparable computational complexity. EmT-B demonstrates higher overall decoding performance across the three datasets with a balance between performance and computational cost.}

\begin{figure}[t]
    \centering
        \subfigure[Connectivity 1]{
    \includegraphics[width=0.45\linewidth]{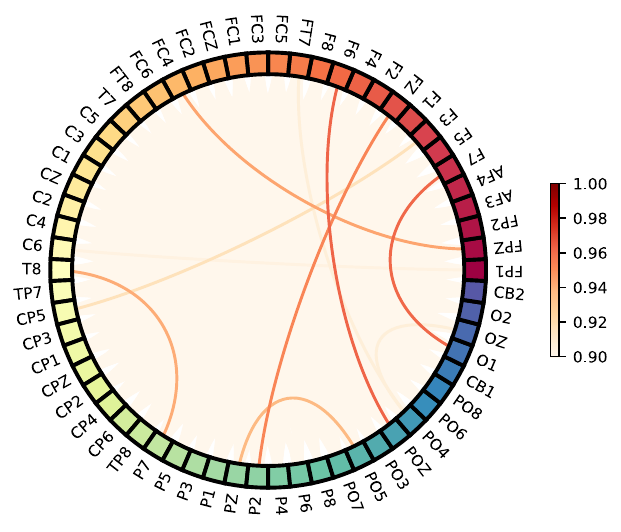}
    }
    \subfigure[Connectivity 2]{
    \includegraphics[width=0.45\linewidth]{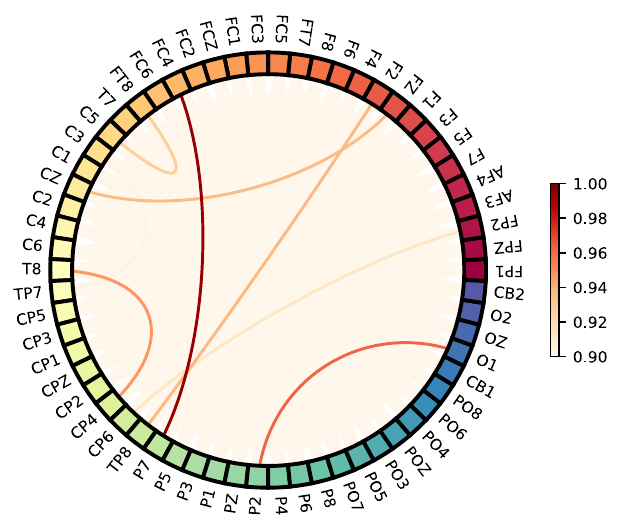}
    }
    \caption{Group-level connectivity of EEG channels for the emotional cognitive process. The two connectivity plots demonstrate that different connectivities of the brain are learned for the emotional task. (a) shows the first learned brain connectivity. (b) shows the second learned brain connectivity. }
    \label{fig:connectivity}
\end{figure}

\subsection{Visualization}
\subsubsection{The learned connections}
Fig.~\ref{fig:connectivity} illustrates the learned connectivity of brain regions on SEED. Two learnable adjacency matrices reveal different connectivity patterns during the emotional cognitive process. Fig.~\ref{fig:connectivity} clearly shows a difference in the learned connectivities under emotion stimulation. In Fig.~ \ref{fig:connectivity} (a), the strongest learned connections are F6-POz, P2-Fz, F7-O1, PO5-Pz, T8-P7, and FC4-FPz. These connections indicate relationships among the frontal, parietal, and temporal areas, which are known to be closely related to mental attention \cite{10.1093/cercor/bhu204}. In Fig.~\ref{fig:connectivity} (b), the important connections learned by EmT are FC4-P7, FT8-C5, Cz-Fz, T8-CP4, CP6-FP2, F2-TP8, and O1-P2. These connections include interactions among the frontal, temporal, and parietal areas, which are known to be related to emotions \cite{gao2021novel, huang2012asymmetric, mickley2009effects}. Additionally, some interactions involve the occipital and parietal areas, which are associated with visual processes. This is expected as videos were used as stimuli in the data collection experiments \cite{zheng2015investigating}. The two distinct patterns in Fig.~\ref{fig:connectivity} reflect EmT's ability to capture multiple cognitive connectivities for the given classification task.

\begin{figure}[t]
    \centering
        \subfigure[Low valance]{
    \includegraphics[width=\linewidth]{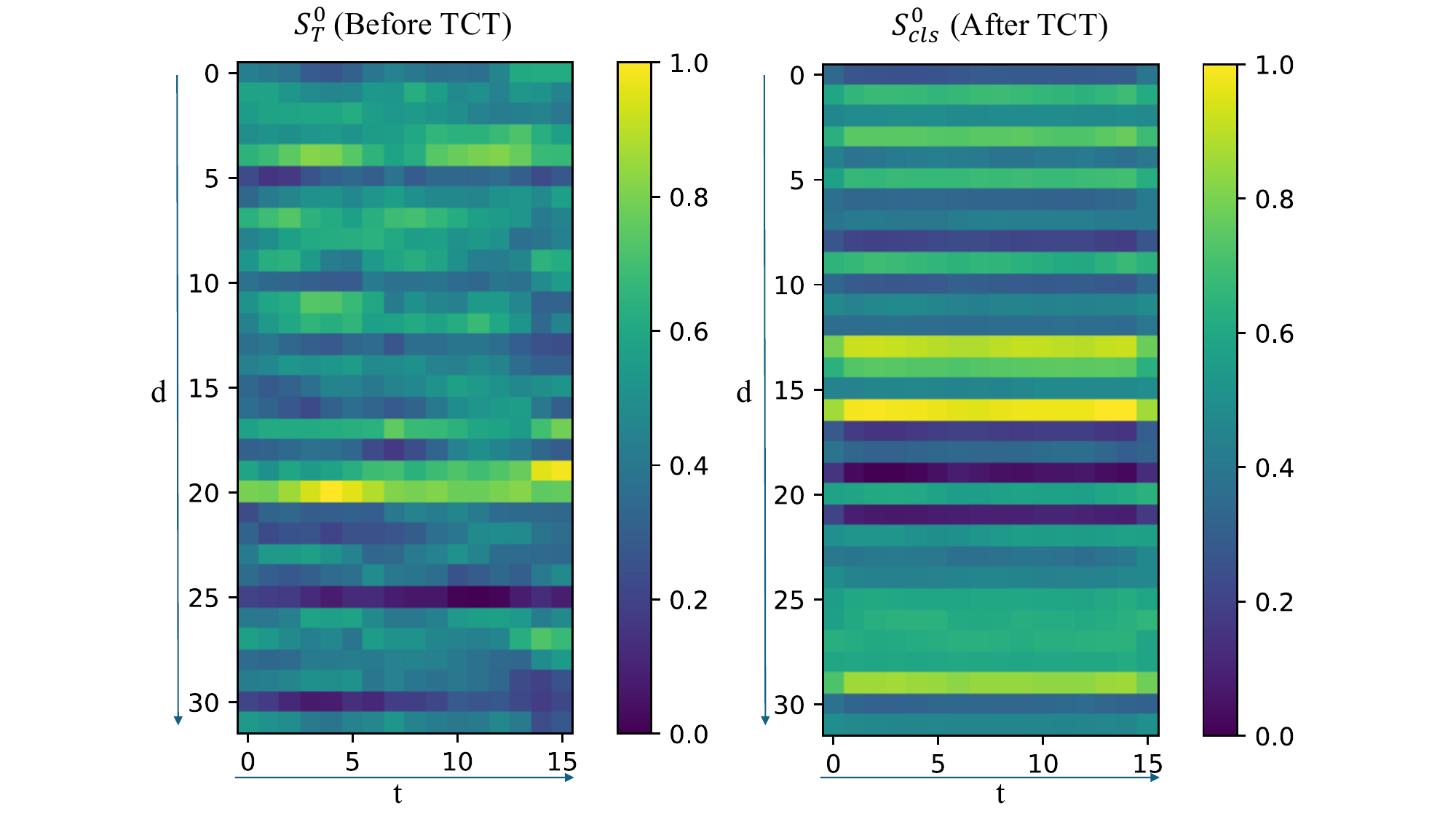}
    }
    \subfigure[High valance]{
    \includegraphics[width=\linewidth]{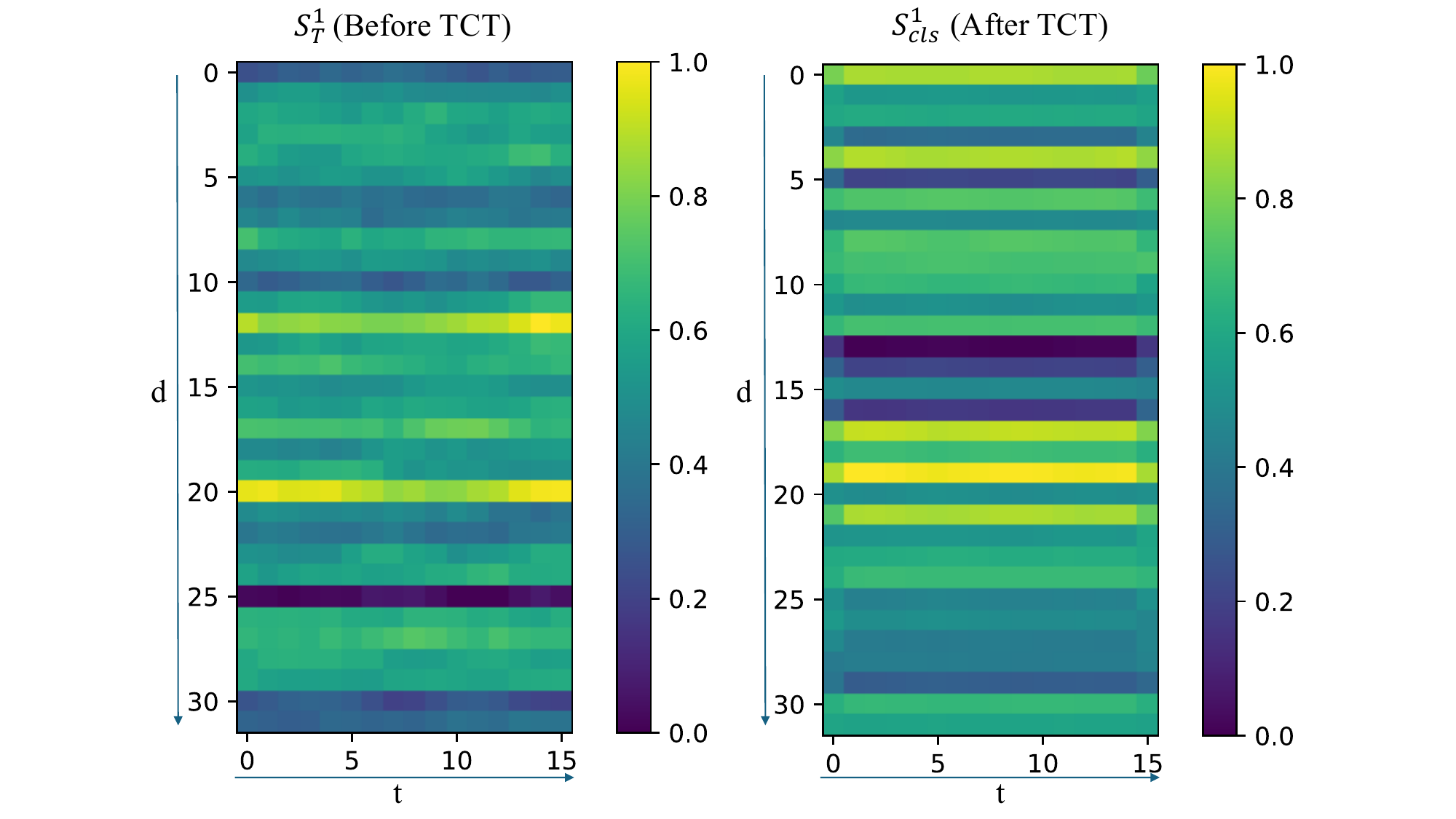}
    }
    \caption{\drop{Visualization of the learned temporal features before and after TCT blocks for the classification task. $d$ represents the dimension of hidden features, and $t$ denotes time. $S_T \in \mathbb{R}^{seq \times d_g}$ and $S_{cls} \in \mathbb{R}^{seq \times d_{head}}$ represent the learned features before and after the TCT blocks, respectively, with superscripts indicating class labels.}}
    \label{fig:feature_clas}
\end{figure}

\begin{figure}[t]
    \centering    
    \includegraphics[width=\linewidth]{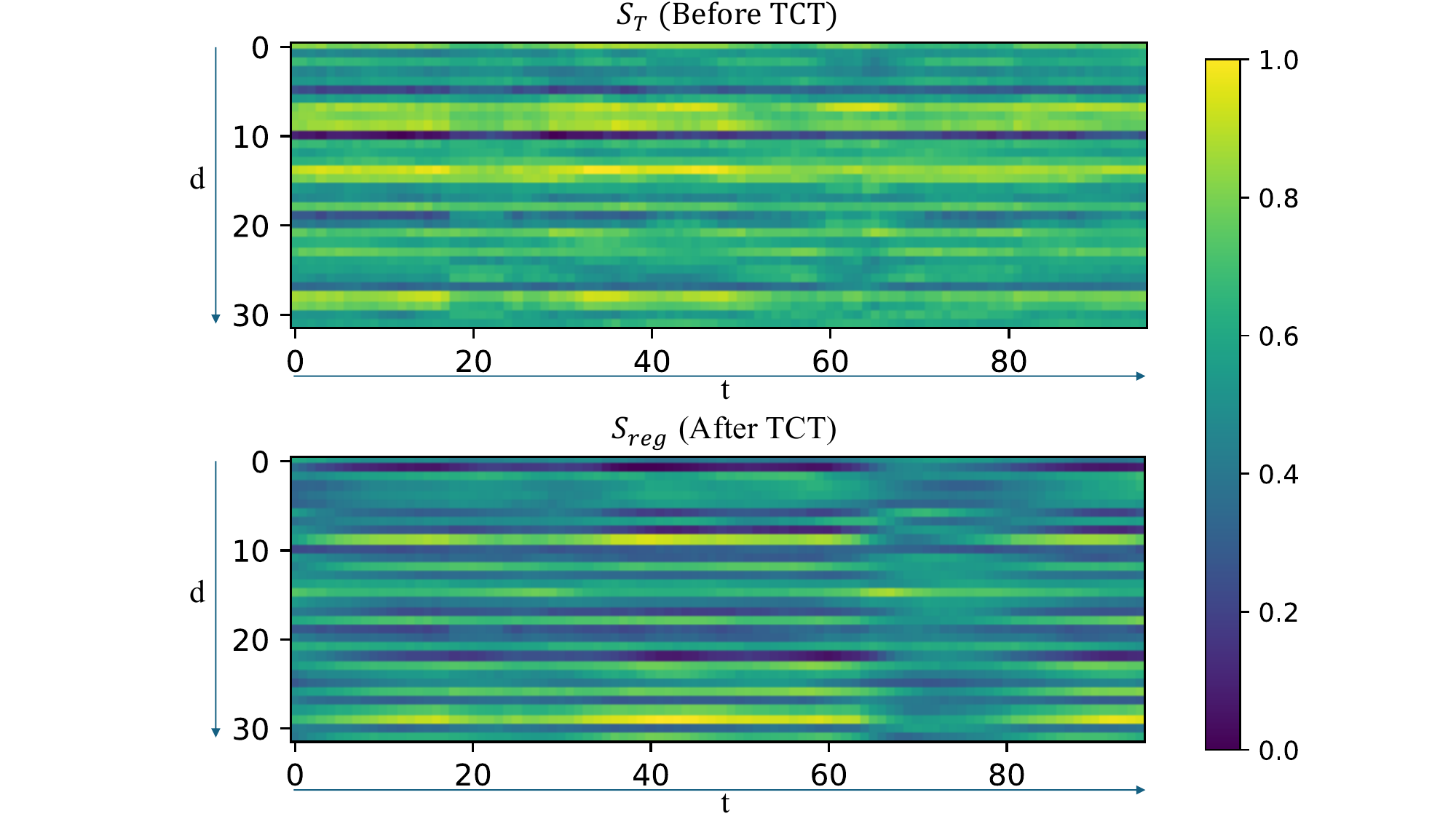}
    \caption{\drop{Visualization of the learned temporal features before and after TCT blocks for the regression task. $d$ represents the dimension of hidden features, and $t$ denotes time. $S_T \in \mathbb{R}^{seq \times d_g}$ and $S_{reg} \in \mathbb{R}^{seq \times d_{head}}$ represent the learned features before and after the TCT blocks.}}
    \label{fig:feature_regr}
\end{figure}

\drop{\subsubsection{The learned temporal contextual information}
We visualized the hidden features before and after the TCT blocks to examine 1) the nature of temporal contextual information and 2) how TCT uses it for classification and regression tasks. For classification, we visualized the averaged features of samples within the same class for each subject, with a representative subject in FACED used to demonstrate the learned temporal contextual information. For regression, we visualized individual sample features, as each has a unique ground truth, and a representative sample is used for illustration purposes.

The visualization results for the classification task are shown in Fig.~\ref{fig:feature_clas}. Here, $d$ represents the dimension of hidden features, and $t$ denotes time. $S_T \in \mathbb{R}^{seq \times d_g}$ and $S_{cls} \in \mathbb{R}^{seq \times d_{head}}$ represent the learned features before and after the TCT blocks, respectively, with superscripts indicating class labels. As seen in Fig.~\ref{fig:feature_clas}, the features change over time before the TCT-Clas blocks, particularly for class 0 (low valence stimuli). After the TCT blocks, the activations become more consistent, likely due to self-attention focusing on parts strongly correlated with the emotional state and the SAT layer smoothing fluctuations by aggregating nearby temporal information. For regression, the visualization (Fig.~\ref{fig:feature_regr}) also shows temporal changes in the feature space, indicating the presence of temporal contextual information. Unlike classification, the regression features are not simply smoothed. Instead, TCT-Regr, using RNNs as token mixers, combines evolving features across different dimensions, resulting in a more complex representation that is well-suited for continuous emotional state prediction. The RNN's ability to capture sequential dependencies allows it to focus on key changes necessary for continuous prediction.

In summary, the changes in features before the TCT blocks confirm the presence of temporal contextual information reflecting the dynamic nature of emotional states. The TCT blocks handle tasks differently: in classification, features are smoothed to enhance separability, while in regression, RNN token mixers preserve temporal variations, enabling continuous emotional predictions.
}

\drop{\subsection{Limitations and Future Work}



Although our model performs better overall across datasets, there are some limitations. First, the model's size and computational complexity can be further optimized. Additionally, while our model improves emotion recognition, the temporal contextual information learned remains difficult to interpret. Future work should design experiments to pinpoint this information. Exploring alternative methods, such as using the Riemannian manifold to capture EEG connection patterns, would also be a promising research direction.
}

\section{Conclusion}

In this paper, we propose a graph-transformer-based model, EmT, for cross-subject EEG emotion recognition. RMPG is proposed to learn multiple connection patterns for different brain cognitive processes under emotional stimulation. TCT is designed to learn temporal contextual information from the temporal EEG graphs. Subject-independent classification and regression tasks are conducted to evaluate EmT and relevant baseline methods. Results on three bench-marking datasets demonstrate EmT shows improvements over the compared baselines. \drop{Our approach is inspired by neuropsychological knowledge and offers a new perspective for improving decoding performance by learning multiple cognition-related graph connections and capturing temporal contextual information. These improvements are supported by enhanced performance and visualizations of the learned connections and temporal features. Additionally, the visualization of the temporal features provides insights into how different token mixers function in classification and regression tasks.}

\section*{Acknowledgment}
This work was supported by the RIE2020 AME Programmatic Fund, Singapore (No. A20G8b0102) and the Agency for Science, Technology and Research (A*STAR) under its MTC Programmatic Funding Scheme (project no. M23L8b0049) Scent Digitalization and Computation (SDC) Programme. 
\ifCLASSOPTIONcaptionsoff
  \newpage
\fi

\bibliographystyle{./Transactions-Bibliography/IEEEtran}
\bibliography{./Transactions-Bibliography/mybib}

\end{document}